\documentclass{article}
\usepackage{ijcai25}

\usepackage{times}
\usepackage{soul}
\usepackage{url}
\usepackage[hidelinks]{hyperref}
\usepackage[utf8]{inputenc}
\usepackage[small]{caption}
\usepackage{graphicx}
\usepackage{amsmath}
\usepackage{amssymb}
\usepackage{amsthm}
\usepackage{booktabs}
\usepackage{algorithm}
\usepackage{algorithmic}
\usepackage{subcaption}
\usepackage[switch]{lineno}

\title{Two-Stage Feature Generation with Transformer and Reinforcement Learning}

\author{
Wanfu Gao$^{1,2}$
\and
Zengyao Man$^{1,2}$\and
Zebin He$^{1,2}$\and
Yuhao Tang$^{1,2}$\and
Jun Gao$^{1,2}$\thanks{Corresponding author}\And
Kunpeng Liu$^{3}$
\affiliations
$^1$College of Computer Science and Technology, Jilin University, China\\
$^2$Key Laboratory of Symbolic Computation and Knowledge Engineering of Ministry of Education, Jilin University, China\\
$^3$Department of Computer Science, Portland State University, Portland, OR 97201 USA\\
\emails
gaowf@jlu.edu.cn,
manzy23@mails.jlu.edu.cn,
hezb2121@mails.jlu.edu.cn,
tangyh2121@mails.jlu.edu.cn,
gaocheng23@mails.jlu.edu.cn,
kunpeng@pdx.edu
}

\begin{document}

\maketitle

\begin{abstract}
Feature generation is a critical step in machine learning, aiming to enhance model performance by capturing complex relationships within the data and generating meaningful new features. Traditional feature generation methods heavily rely on domain expertise and manual intervention, making the process labor-intensive and challenging to adapt to different scenarios. Although automated feature generation techniques address these issues to some extent, they often face challenges such as feature redundancy, inefficiency in feature space exploration, and limited adaptability to diverse datasets and tasks.
To address these problems, we propose a Two-Stage Feature Generation (TSFG) framework, which integrates a Transformer-based encoder-decoder architecture with Proximal Policy Optimization (PPO). The encoder-decoder model in TSFG leverages the Transformer’s self-attention mechanism to efficiently represent and transform features, capturing complex dependencies within the data. PPO further enhances TSFG by dynamically adjusting the feature generation strategy based on task-specific feedback, optimizing the process for improved performance and adaptability.
TSFG dynamically generates high-quality feature sets, significantly improving the predictive performance of machine learning models. Experimental results demonstrate that TSFG outperforms existing state-of-the-art methods in terms of feature quality and adaptability\footnote{The code is available at https://github.com/fassat/TSFG}. 
\end{abstract}

\begin{figure}[t]
\centering
\includegraphics[width=0.95\columnwidth]{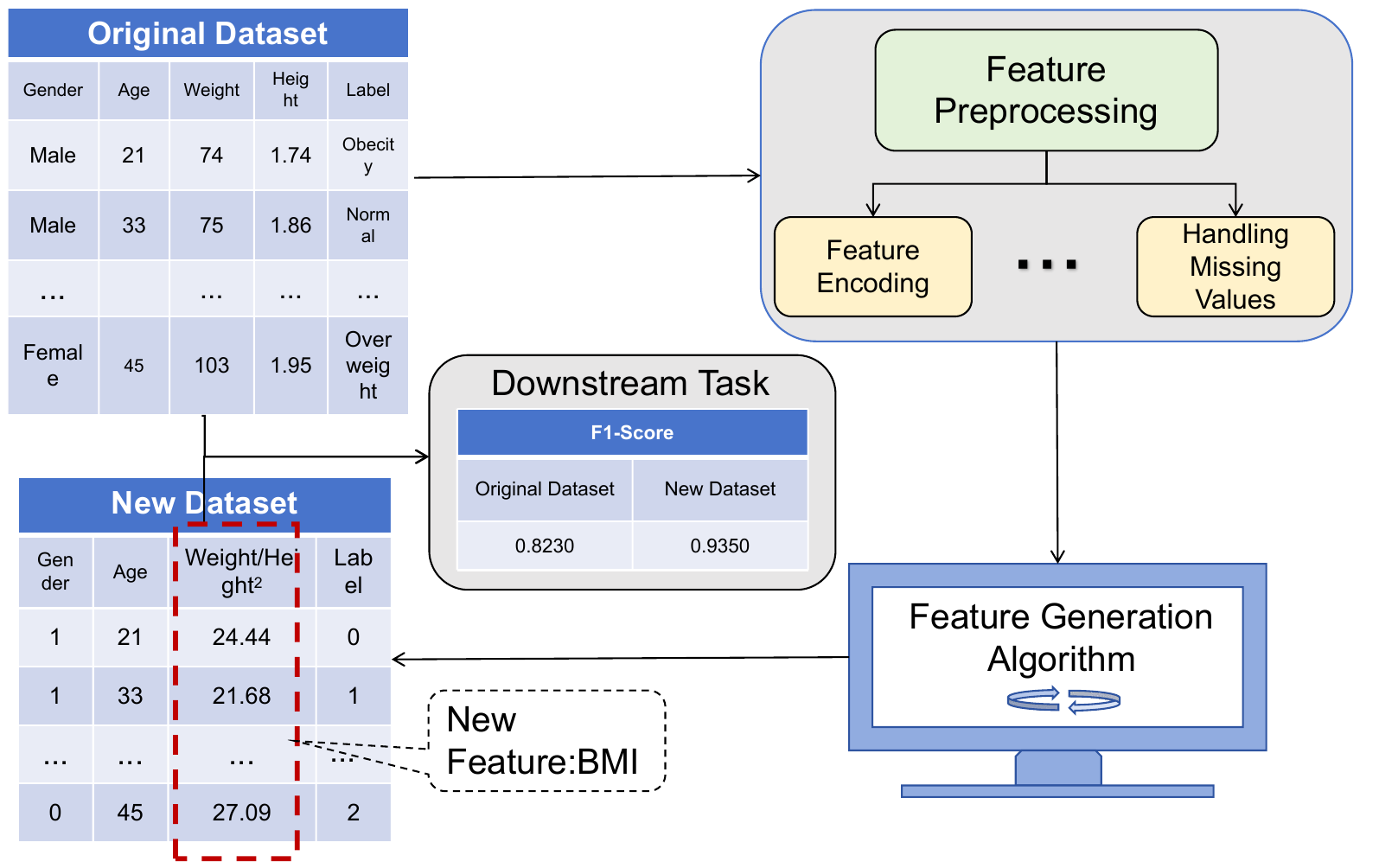}
\caption{The original dataset is preprocessed and then fed into the feature generation algorithm. This algorithm creates new features and produces an enhanced dataset. The enhanced dataset is used for downstream tasks. Results show that the F1 score of the enhanced dataset is better than the original dataset.}
\label{fig_1}
\end{figure}

\section{Introduction}

Feature generation is a critical aspect of the construction of high-performance machine learning models. High-quality features can enhance the robustness, generalization capability, and interpretability of models \cite{feature,dong2018feature}. For instance, as Figure 1 shows, combining weight and height features into a Body Mass Index (BMI) \cite{bmi} feature can significantly improve a model's ability to predict health conditions. As the scale and complexity of data increase, traditional feature engineering methods often struggle to uncover the complex patterns and underlying relationships within the data. Consequently, automatic feature generation techniques, especially those based on deep learning, have become a focal point of research \cite{automl}.  Recently, reinforcement learning has also provided new ideas for feature generation \cite{Liu2019AFSE,wang2022group}. Despite their ability to extract complex features, deep learning methods still face challenges in adapting to different tasks and achieving efficient optimization \cite{lecun2015deep}. This study introduces a two-stage feature generation framework that integrates pre-training with reinforcement learning-based fine-tuning to achieve more efficient and precise feature generation.

The novelty of this framework lies in its staged training method, which enables stepwise optimization of feature generation through pre-training and reinforcement learning fine-tuning. In the first stage of pre-training, we employ an Encoder-Decoder architecture \cite{sequence}, optimizing the model using cross-entropy loss or mean squared error loss. The encoder extracts latent information from the raw dataset, while the decoder reconstructs this information into logically coherent sequences of feature combinations. For example, initial features might appear as \([ V1, \text{EOS}, V2, \text{EOS}, V3, \text{STOP} ]\), and during decoding, more intricate feature combinations like \([ +V1, +V2, \text{EOS}, -V3, \times V1, \text{STOP} ]\) are generated. The goal of this stage is to build a stable initial model for feature generation.

Upon completing the first stage of pre-training, the second stage incorporates the Proximal Policy Optimization (PPO) algorithm \cite{PPO2017} from reinforcement learning to fine-tune the model. This phase dynamically adjusts the model weights through a reward mechanism tied to specific downstream tasks, ensuring that the generated features better align with the requirements of these tasks. Specifically, the reward value of the generated feature combinations will be calculated. Based on the reward signal, the model optimizes its generation strategy, progressively learning high-quality feature representations. The PPO algorithm excels at balancing the magnitude of updates, preventing issues such as gradient explosion during the training process.

This two-stage training strategy offers several advantages. In the first stage, the Encoder-Decoder structure generates features step by step and computes the loss at each step, ensuring the stability and consistency of the generated features. In the second stage, the introduction of the PPO algorithm endows the feature generation process with dynamic optimization capabilities, enhancing task adaptability. 

In summary, this paper introduces a novel feature generation framework, which leverages an encoder-decoder architecture integrated with Proximal Policy Optimization (PPO). The proposed framework enhances feature generation by dynamically adjusting the optimization strategy and improving adaptability across different tasks. Using a two-stage training process, we address the challenges of feature redundancy and inefficient exploration in traditional methods. Our main contributions include:

\begin{itemize}
\item Encoder-Decoder Architecture for Efficient Feature Generation: We propose an encoder-decoder architecture that effectively models the feature generation process, allowing for the extraction of meaningful latent representations from raw data. This ensures high-quality feature combinations for downstream tasks.

\item Two-Stage Training with PPO Fine-Tuning: We introduce a two-stage method where pre-training initializes the model, and reinforcement learning fine-tuning optimizes feature generation based on specific task requirements. This allows the framework to dynamically adjust its strategy and enhance task performance.

\item Enhanced Adaptability and Efficiency: By incorporating reinforcement learning and task-specific feedback, our framework can efficiently generate features tailored to diverse datasets and models, offering better performance and computational efficiency compared to existing methods.
\end{itemize}

\begin{figure*}[t]
\centering
\includegraphics[width=0.8\textwidth]{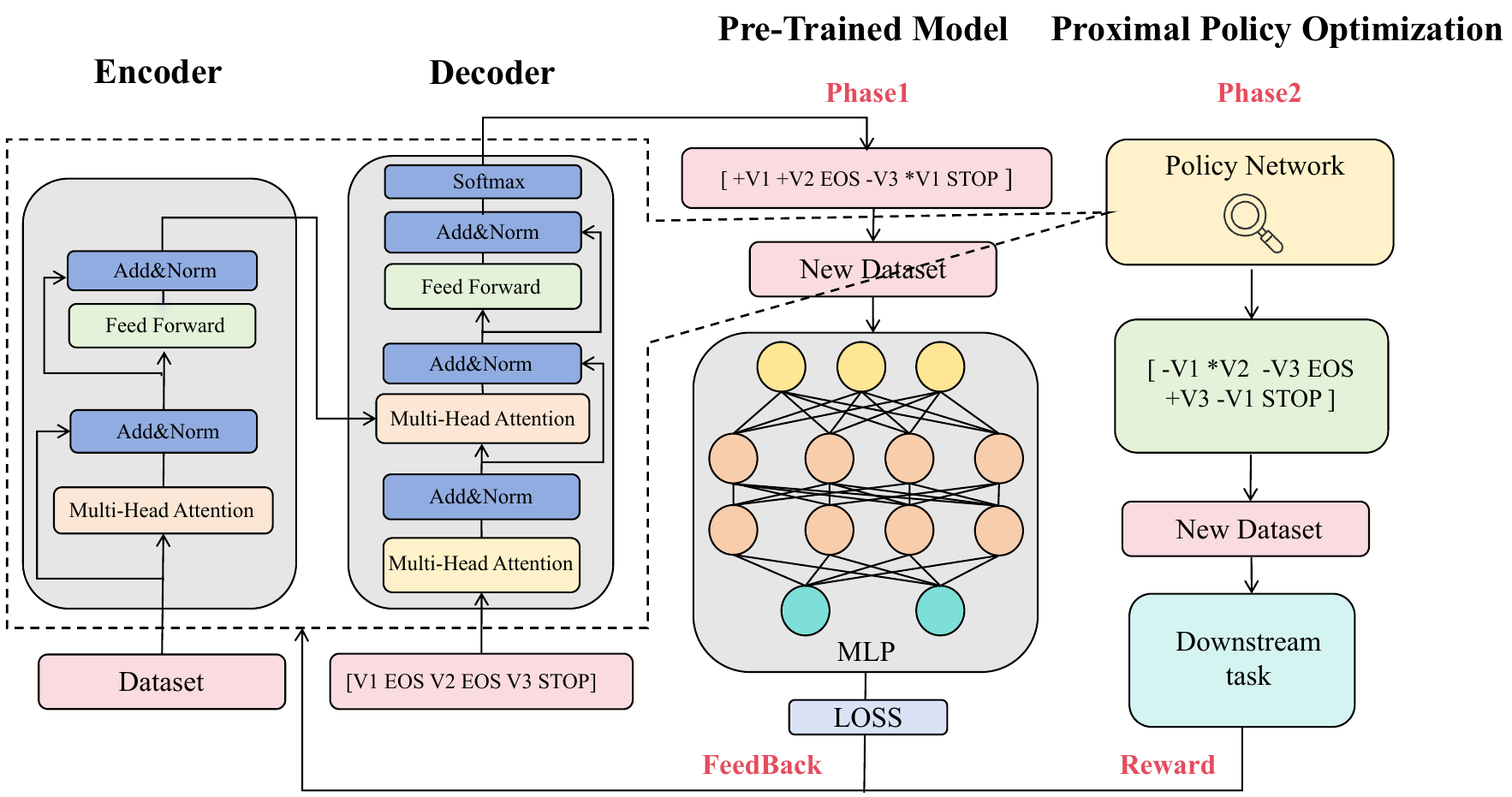} 
\caption{Overview of TSFG. In the first phase, the encoder-decoder model is optimized by calculating the loss. In the second phase, the encoder-decoder model serves as a policy network using Proximal Policy Optimization (PPO) to fine-tune the model based on reward feedback. This optimizes feature generation to adapt to specific downstream tasks, ensuring that feature generation can efficiently adapt to a variety of task requirements.}
\label{fig_framework}
\end{figure*}

\section{Related Work}

Automated feature generation plays a critical role in enhancing the performance of machine learning models \cite{domingos2012few}. Common methods in this field can generally be divided into expansion-reduction methods and search-based methods, both aiming to generate meaningful and high-quality features for predictive modeling.

Expansion-reduction methods generate a large pool of candidate features through various transformations and then reduce redundancy by selecting the most relevant subset. For example, ExploreKit \cite{katz2016explorekit} follows a three-step workflow: candidate feature generation, ranking, and evaluation. It applies transformation functions to the entire dataset and evaluates the performance of the predictive model to select the most valuable features. Similarly, FEADIS \cite{dor2012strengthening} generates new features by randomly combining original features with mathematical functions, while Autofeat \cite{horn2020autofeat} expands the feature space using nonlinear transformations before selecting a small subset of features for inclusion. OpenFE \cite{openfe2023} is an efficient expansion-reduction method that combines FeatureBoost and a two-stage pruning algorithm, enabling it to rapidly and accurately identify useful new features on large-scale datasets.

Despite their contributions, these methods often face challenges related to the computational burden and redundancy caused by the exponential growth of candidate features. As the number of features increases, selecting truly relevant ones becomes increasingly difficult, This prompts researchers to develop alternative approaches that enhance representation completeness \cite{wang2021generative}.

Search-based methods adopt a more targeted method by using search algorithms to explore potential feature transformations and combinations. These methods have shown potential in overcoming the limitations of expansion-reduction techniques. For instance, GRFG \cite{wang2022group} utilizes group reinforcement learning to select operations for groups of features and performs operations between them, effectively managing feature dependencies. TransGraph \cite{khurana2018feature} uses a transformation graph and Q-learning algorithm to generate higher-order features. DIFER \cite{difer2022} introduces a feature optimizer within an encoder-predictor-decoder structure that maps features to a continuous vector space, optimizes embeddings along the gradient direction induced by the predictor, and recovers better features from the optimized embeddings via the decoder.

In addition, some studies on multi-label feature selection have proposed relevance metrics and feature decomposition strategies to ensure task-specific feature quality while reducing redundancy \cite{cond-weight,rel-var,shared-feat}.

In summary, while expansion-reduction methods can create a vast array of potential features, search-based methods offer a more strategic exploration of the feature space \cite{komer2014hyperopt}, potentially leading to more effective and efficient feature engineering.

\section{The Proposed Method}

\subsection{Problem Settings}
Given a dataset $\mathcal{D} = \{ \mathcal{F}, y \}$, where $\mathcal{F}$ represents the original feature set and $y$ represents the target label set. The feature set $\mathcal{F} = \{V_1, V_2, \ldots, V_N\}$ consists of $N$ discrete features. 

We aim to find the optimal feature set $\mathcal{F}^*$ that maximizes the performance metric of downstream tasks, defined as:

\begin{equation}
\mathcal{F}^* = \underset{\mathcal{F}'}{\arg\max} \; M_A(\mathcal{F}', y),
\end{equation}
where $A$ represents a downstream machine learning model (e.g., random forests, SVM, or neural networks), and $M$ denotes the evaluation metric, such as accuracy, \textit{F1-score}, or mean squared error.

For continuous features, mathematical operations include ``absolute value", ``square", ``inverse", ``logarithm", ``square root", ``cube", ``addition", ``subtraction", ``multiplication", and ``division". For discrete features, mathematical operations include ``cross" \cite{luo2019autocross} and ``addition".

\subsection{Overall Framework}

The framework combines reinforcement learning, an encoder-decoder architecture, and a feature transformation mechanism, aiming to generate optimized features to address challenges such as avoiding redundant feature generation and optimizing feature transformation sequences. The workflow is based on an encoder-decoder architecture as the main structure, a pre-training phase for initialization, and a reinforcement learning phase for fine-tuning.

At the core of the framework is an encoder-decoder architecture that can efficiently explore the feature space and generate new features. We select the Transformer architecture \cite{trans} as the encoder-decoder model due to its superior capability in handling sequential problems \cite{raffel2020exploring}. Its self-attention mechanism effectively captures long-range dependencies and complex patterns within the data \cite{bahdanau2014neural}. The encoder processes the input dataset, mapping it into a latent representation that captures the structure and interdependencies of the original features. This latent representation serves as the foundation from which the decoder generates sequences of feature transformations. 

The decoder is responsible for generating feature transformation sequences, which are produced step-by-step through an autoregressive mechanism \cite{graves2013generating}. The output of each step depends on the result of the previous step, ensuring that the feature transformation process proceeds coherently and effectively captures the dependencies among features. To avoid generating invalid features, operations are combined with features to form a unified token. For example, operations like \( +V1 \), \( -V2 \), etc., are treated as tokens. The sequence output by the decoder is similar to \([ +V1, +V2, \text{EOS}, -V3, \times V1, \text{STOP} ]\), where \(\text{EOS}\) represents the end-of-feature token and \(\text{STOP}\) represents the end-of-sequence token. In this way, we generate two new features: \( V1 + V2 \) and \( -V3 \times V1 \), thereby expanding the feature set and improving the data representation for downstream machine learning tasks.

The pre-training phase focuses on initializing the encoder-decoder model. The encoder extracts latent representations of the dataset, and the decoder generates feature transformation sequences based on these representations. These transformations generate an augmented feature set, which is optimized by evaluating its performance on a pre-trained MLP model. The optimization objective is to minimize the cross-entropy loss or mean squared error loss, thereby enhancing the quality of feature generation and its adaptability to tasks.

To encourage exploration during training, a sampling mechanism with temperature scaling is adopted \cite{guo2017calibration}. This mechanism adjusts the sampling probabilities of certain transformations, enabling the model to explore a wide range of feature combinations. The pre-trained encoder-decoder provides a solid foundation for the reinforcement learning phase, accelerating convergence and enhancing performance.

After pre-training, the model is fine-tuned using the Proximal Policy Optimization (PPO) algorithm. This phase focuses on iteratively improving the feature generation process to maximize cumulative rewards. The reward function is based on improvements in the performance of the downstream task, such as increased classification accuracy or F1 score.
In each iteration of the PPO phase, similar to the pre-training stage, the encoder generates a latent representation of the dataset, and the decoder proposes candidate feature transformations based on this representation. These transformations are evaluated using external performance metrics, and a reward is calculated for the generated sequence. The PPO algorithm adjusts the parameters of the encoder-decoder to maximize the expected reward, balancing exploration and exploitation.

The PPO fine-tuning enables the framework to dynamically adapt to the data, generating task-optimized feature sets. This method not only ensures that the generated features enhance performance but also maintains their interpretability and relevance to the problem domain.

\subsection{Pre-Training Phase for Encoder-Decoder Model}

The goal of the pre-training phase is to enable the encoder-decoder model to learn an initial strategy. A new feature set is generated by the transformation sequence, which is then processed by the pre-trained MLP to compute predictions and calculate the prediction loss based on the true labels. By minimizing the prediction loss between the generated transformation sequences and the true labels, the model learns how to enhance the representations of features. The encoder-decoder is implemented based on the Transformer architecture.

\textbf{Input:} The input for the pre-training phase includes a dataset \( D_{\text{train}} = \{ F, y \} \), where \( F \) represents the original feature set, and \( y \) is the target variable. Additionally, a validation dataset \( D_{\text{val}} \) is used to evaluate the model's performance and generalization ability during training.

\textbf{Output:} The output is a pre-trained encoder-decoder model capable of generating feature transformation sequences for creating new features. 

The encoder maps the input features \( F \) into a latent space \( \mathcal{Z} \), where the latent representation \( z \in \mathcal{Z} \) captures the structural relationships among the features. This ensures that the decoder has a comprehensive understanding of the data, allowing it to extract hierarchical feature representations. Formally, this process is expressed as:

\begin{equation}
z = \text{Encoder}(F; \theta_e),
\end{equation}
where \( \theta_e \) represents the parameters of the encoder. The latent space \( Z \) effectively captures the complex relationships between features, providing a compact yet informative representation for subsequent transformation.

The decoder then receives the latent representation \( z \) and predicts the transformation operation 
\( a \), which is applied to the features. This process can be formalized as:

\begin{equation}
a = \text{Decoder}(z; \theta_d),
\end{equation}
where \( \theta_d \) represents the parameters of the decoder.

Furthermore, the decoder uses a temperature parameter \( T \) to adjust the probability distribution of the output sequence \cite{Graves2013GeneratingSequences}. 

\begin{equation}
P(a , z) = \frac{\exp(a / T)}{\sum_{i} \exp(a_i / T)},
\end{equation}
where \( T \) is the temperature parameter, \( a_i \) represents one of all possible actions. The temperature parameter controls the smoothness of the output probability distribution. When the temperature \( T \) is high, the generated features become more random, leading to a more exploratory method. Conversely, when \( T \) is low, the model tends to select features with higher probabilities, resulting in more deterministic outputs.

In this process, the decoder generates new feature transformation operations. These operations are applied to the original training dataset \( D_{\text{train}} \), creating a new dataset \( D_{\text{train}}^{\text{new}} \). The dataset \( D_{\text{train}}^{\text{new}} \) is then fed into a pre-trained MLP, which outputs the predicted probability distribution for each sample's class. Finally, the loss between these predicted probabilities and the true labels is calculated to evaluate the quality of the generated feature transformations. The parameters of the encoder-decoder model are adjusted through backpropagation to minimize the loss function. The framework inputs the data into the MLP to compute the loss after generating a complete feature (i.e., when the EOS token is generated), rather than waiting to generate the entire transformation sequence. This step-by-step evaluation provides timely feedback on the effectiveness of each generated feature, thereby promoting faster convergence during training and further improving the quality of the generated features.

For classification tasks, the cross-entropy loss is defined as:

\begin{equation}
L_{\text{classification}} = -\frac{1}{N} \sum_{i=1}^{N} \left[ y_i \log(\hat{y}_i) + (1 - y_i) \log(1 - \hat{y}_i) \right],
\end{equation}
where \(N\) is the number of training samples. \(y_i\) is the true label of the \(i\)-th training sample. \(\hat{y}_i\) is the predicted probability for the \(i\)-th sample, obtained by the pre-trained MLP after applying the current generated features to the training dataset \(D_{\text{train}}\).

For regression tasks, the mean squared error (MSE) loss is defined as:

\begin{equation}
L_{\text{regression}} = \frac{1}{N} \sum_{j=1}^{N} \left( \hat{y}_j - y_j \right)^2,
\end{equation}
where \(N\) is the number of training samples. \(\hat{y}_j\) is the predicted value of the \(j\)-th feature. \(y_j\) is the true value of the target feature for the \(j\)-th sample.

\subsection{Proximal Policy Optimization (PPO)}

The Proximal Policy Optimization (PPO) fine-tuning phase is designed to refine the feature generation process by leveraging reinforcement learning. While the pre-training phase provides a strong initialization for the encoder-decoder architecture, the PPO phase ensures that the model continues to optimize feature transformations based on feedback from downstream tasks. This phase is critical for dynamically adapting to specific datasets and tasks. The final new dataset is generated at this stage.

The objective of the PPO fine-tuning phase is to iteratively improve the feature generation process by maximizing a reward function that reflects the performance improvement of downstream tasks. 
Unlike the pre-training phase, where training is performed after generating each feature, in the PPO phase, the model parameters are optimized based on feedback from downstream tasks after generating the complete transformation sequence. This method comprehensively evaluates the overall performance of the complete feature set and the synergistic relationships among features, ensuring that the optimization direction aligns more closely with the requirements of downstream tasks, thereby producing higher-quality feature sets with greater global coherence.

The feature generation process is modeled as a Markov Decision Process (MDP). The encoder-decoder serves as the policy network, with the state, action, and reward defined as follows:

\textbf{State} ($s$): The current state represents the feature set.

\textbf{Action} ($a$): An action corresponds to a transformation operation selected by the decoder. For example, \(+V_1\), \(-V_2\), \(*V_2\).

\textbf{Reward} ($r$): The reward is the performance improvement of the new dataset in the downstream task. The reward can be expressed as:
\begin{equation}
r = \Delta \text{Metric} = \text{Metric}(\mathcal{D}_{\text{new}}) - \text{Metric}(\mathcal{D}),
\end{equation}
where $\mathcal{D}_{\text{new}}$ is the dataset with the newly generated feature set.

\textbf{Policy} ($\pi$): The policy determines which action to apply, given the current state. PPO is used to iteratively optimize the policy to maximize cumulative rewards.

The PPO fine-tuning workflow begins with initializing the encoder-decoder model using the weights obtained during the pre-training phase. This ensures that the model starts fine-tuning with a well policy for feature generation.

At each iteration, the encoder generates latent representations of the current dataset, which are passed to the decoder. The decoder samples candidate transformation sequences based on the current policy \(\pi_{\theta}(a|s)\). The sequence is applied to the original dataset to generate a new dataset. The updated dataset is evaluated using a downstream machine learning model (e.g., a classifier or regressor). The model calculates the sequence's reward based on the performance improvement achieved by the new dataset.

Using the computed rewards, PPO adjusts the policy parameters \(\theta\) to maximize the expected cumulative reward. The policy objective is defined as:
\begin{equation}
L_{\text{PPO}}(\theta) = \mathbb{E}_t \left[ \min \left( r_t(\theta) \hat{A}_t, \text{clip}(r_t(\theta), 1 - \epsilon, 1 + \epsilon) \hat{A}_t \right) \right],
\end{equation}
where \(r_t(\theta) = \frac{\pi_{\theta}(a_t|s_t)}{\pi_{\theta_{\text{old}}}(a_t|s_t)}\) is the probability ratio, \(\hat{A}_t\) is the advantage function, and \(\epsilon\) is a clipping parameter to limit the magnitude of policy updates.
To encourage the exploration of diverse transformations, an entropy regularization term is added to the PPO objective \cite{Zhang2023EntropyRegularization}. This term prevents the policy from converging prematurely to suboptimal solutions by promoting randomness in action selection.
The process of sampling, evaluation, and policy updates is repeated for a predefined number of iterations or until convergence criteria are met.

The PPO fine-tuning phase offers several advantages. Stability is achieved through PPO's clipping mechanism, which ensures stable training by limiting large updates to the policy. The reward-driven method allows the model to dynamically adapt to specific datasets and tasks.
\begin{algorithm}[t]
\caption{Encoder-Decoder Training with PPO Fine-Tuning}
\begin{algorithmic}[1]
\STATE \textbf{Input:} Dataset \( \mathcal{D} = \{ \mathcal{F}, y \} \), where \( \mathcal{F} = \{V_1, V_2, \ldots, V_N\} \) is the feature set and \( y \) is the target label.
\STATE \textbf{Output:} Optimized feature set \( \mathcal{F}^* \).
\STATE \textbf{Pre-Training Phase:}
\STATE \quad Train the encoder-decoder model using a supervised loss function (cross-entropy or MSE).
\STATE \quad For each feature:
\STATE \quad \quad Compute loss \( L_{\text{classification}} \) or \( L_{\text{regression}} \).
\STATE \quad \quad Backpropagate and update model parameters.
\STATE \quad Store the best model with minimal validation loss as the initialization for PPO fine-tuning.

\STATE \textbf{PPO Fine-Tuning Phase:}
\STATE For each iteration \( t = 1 \) to \( T_{\text{max}} \):

\STATE \quad Sample transformation operations \( a \) from the policy \( \pi_{\theta}(a|s) \).
\STATE \quad Apply transformations to \( \mathcal{D} \) to generate a new feature set \( \mathcal{D}^{\text{new}} \).
\STATE \quad Evaluate the performance of \( \mathcal{D}^{\text{new}} \) on downstream task using a performance metric (e.g., accuracy, \textit{F1-score}).
\STATE \quad Compute reward \( r \) based on the performance improvement:
\[
r =  \text{Metric}(\mathcal{D}^{\text{new}}) - \text{Metric}(\mathcal{D})
\]

\STATE \quad Compute PPO objective:
{\small
\begin{equation*}
L_{\text{PPO}}(\theta) = \mathbb{E}_t \left[ \min \left( r_t(\theta) \hat{A}_t, \text{clip}(r_t(\theta), 1 - \epsilon, 1 + \epsilon) \hat{A}_t \right) \right]
\end{equation*}
}

\STATE \textbf{End for PPO Iterations}

\end{algorithmic}
\label{alg}
\end{algorithm}


\begin{table*}[ht]
\centering
\renewcommand{\arraystretch}{1.5} 

\resizebox{\textwidth}{!}{ 
\begin{tabular}{c|cccccc|cccccc|cccccc}
\hline
\textbf{Datasets} & \multicolumn{6}{c|}{\textbf{ACC}}&\multicolumn{6}{c|}{\textbf{\textit{F1-Score}}} & \multicolumn{6}{c}{\textbf{Precision}} \\ \hline
classification & Base & GRFG & DFS & OpenFE & DIFER & TSFG & Base & GRFG & DFS & OpenFE & DIFER & TSFG & Base & GRFG & DFS & OpenFE & DIFER & TSFG \\ \hline
australian &0.8613 &\underline{0.8759} &0.8467 &0.854 &0.8467 &\textbf{0.8832} 
&0.8613 &\underline{0.8757} &0.8466 &0.854 &0.8467 &\textbf{0.8832} &0.8613 &\underline{0.8765} &0.8472 &0.8544 &0.8475 &\textbf{0.8832}\\ 

credit\_g &0.71 &0.775 &0.725 &0.71 & \underline{0.78}&\textbf{0.785} 
&0.698 &0.7756 &0.7257 &0.7116 &\underline{0.7773} &\textbf{0.7817} 
&0.6913 &\underline{0.7763} &0.7266 &0.7133 & 0.7754&\textbf{0.7795}  \\ 

diabetes &0.719 &0.719 &0.7059 &\underline{0.732} &0.7273 &\textbf{0.7582} 
&0.7062 &0.7115 &0.6954 &\underline{0.7233} &0.7179 &\textbf{0.7503} 
&0.7182 &0.7151 &0.7021 &\underline{0.73} &0.7182 & \textbf{0.7583} \\ 

f5 & 0.795&0.76 &0.7725 &\underline{0.805} &0.7925 &\textbf{0.825} 
&0.7943 &0.7602 &0.7719 & \underline{0.8034}&0.7913 &\textbf{0.824} 
& 0.7956&0.7623 &0.7726 &\textbf{0.888} &0.7945 &\underline{0.8273}  \\ 

hepatitis & 0.8387&0.8387 &0.871 &0.8387 &\textbf{0.9355} &\underline{0.871}
&0.8214 &0.8215 &0.8628 &0.8342 &\textbf{0.9355} &\underline{0.8762} 
&0.8289 &0.8289 &0.8655 &0.8318 &\textbf{0.9355} &\underline{0.8895}  \\ 

ionosphere &\underline{0.9143} &0.9 & \textbf{0.9286}&\underline{0.9143} &\underline{0.9143} & \textbf{0.9286}
&0.9125 &0.8986 &\underline{0.9266} &0.9114 &0.9126 & \textbf{0.9267}
&0.9176 &0.9008 &\textbf{0.9359} & \underline{0.9246}& 0.9176&\textbf{0.9359} \\ 

NPHA &\underline{0.4577} & 0.3803&0.3943 & 0.4225& 0.4155& \textbf{0.4718}
&\underline{0.4465} & 0.3914&0.3898 &0.4199 &0.4003 &\textbf{0.4599} 
&\underline{0.4418} &0.4091 &0.3859 &0.418 &0.3896 &\textbf{0.4534} \\ 

PimaIndian &0.719 &0.7190 & 0.7059& \underline{0.732}& 0.7272&\textbf{0.7321}
&0.7062 &0.7115 &0.6945 & \textbf{0.7233}&\underline{0.7179} & 0.7111
&0.7182 &0.7151 &0.7021 & \underline{0.73}&0.7182 &\textbf{0.7461} \\ 

seismic &0.9167 & 0.9176& \underline{0.9205}& 0.9186& 0.9201& \textbf{0.9205}& 
0.8854& 0.8822& \underline{0.8936}&0.8896 &0.8896 &\textbf{0.8961} 
& 0.8793& 0.8723& 0.8965& 0.889& \textbf{0.9076}&\underline{0.8968} \\
\hline

& \multicolumn{6}{c|}{\textbf{$1\!-\!\mathrm{RAE}$}} & \multicolumn{6}{c|}{\textbf{$R^2$}} & & & & & & \\
\cline{1-7} \cline{8-13}  
regression & Base & GRFG & DFS & OpenFE & DIFER & TSFG & Base & GRFG & DFS & OpenFE & DIFER & TSFG &  &  &  &  &  &  \\ 
\cline{1-7} \cline{8-13}
Openml\_582 & \underline{0.6551}& 0.5644&0.54 &0.6117 & 0.6128&\textbf{0.6612} 
& \underline{0.849}& 0.7946&0.7735 &0.8394 &0.8147 &\textbf{0.8556} & & & & & & \\ 
Openml\_595 &0.5724 &0.4327 &0.5585 &0.6067 &\textbf{0.6193} & \underline{0.6162}
&0.7636 &0.6825 &0.8077 &\underline{0.8385} & \textbf{0.8444}& 0.8114& & & & & & \\ 
Openml\_637 &0.534 &0.5407 &0.5303 &0.4631 &\underline{0.5409} & \textbf{0.5489} 
&0.7326 &\underline{0.7708} & \textbf{0.7709}& 0.6718& 0.7404& 0.7444& & & & & & \\ 
Openml\_639 & 0 & \underline{0.138} & 0 & 0 & 0 & \textbf{0.1981} 
&0 &\underline{0.1104} & 0& 0& 0& \textbf{0.1926}& & & & & & \\ 
\hline
\end{tabular}
}

\caption{Overall Performance. In this table, the best and second-best results are highlighted in bold and underlined fonts respectively. We evaluate classification tasks using \textit{F1-score}, accuracy, and precision, and regression tasks using 1-RAE and $R^2$. The higher the value, the better the quality of the transformed feature set.}
\label{tab:1} 
\end{table*}


\begin{table}[ht]
\centering

\begin{tabular}{lccc} 
\toprule
\textbf{Datasets} &  \textbf{C/R} & \textbf{\#Samples} & \textbf{\#Features} \\
\midrule
australian   & C & 690  & 14 \\
credit\_g    & C & 1000 & 21 \\
diabetes       & C & 768  & 8  \\
f5             & C & 267  & 44 \\
hepatitis    & C & 155  & 19 \\
ionosphere   & C & 351  & 34 \\
NPHA         & C & 714  & 14 \\
PimaIndian   & C & 768  & 8  \\
seismic        & C & 210  & 8  \\
Openml\_582    & R & 500 & 25 \\
Openml\_595    & R & 1000 & 10 \\
Openml\_637    & R & 500  & 50 \\
Openml\_639   & R & 100  & 25 \\
\bottomrule
\end{tabular}
\caption{Dataset Information.} 
\label{tab:2} 
\end{table}

\section{Experiments}

\subsection{Experimental Setup}
\paragraph{Data Description.} We conducte experiments on 13 datasets from UCI \cite{public2024}, Kaggle \cite{howard2024kaggle}, and OpenML \cite{public2024openml}, LibSVM \cite{lin2024libsvm}, comprising 9 classification tasks and 4 regression tasks. Table \ref{tab:2} shows the statistics of the data.

\paragraph{Evaluation Metrics.} For classification tasks, we use \textit{F1-score}, accuracy, and precision to evaluate the dataset. For regression tasks, we use 1-relative absolute error (1-RAE)  and $R^2$ to evaluate the dataset.
The equation for 1-RAE is as follows:
\begin{equation}
1-RAE = 1- \frac{\sum_{i=1}^n \lvert y_i - y_i^*\rvert}{\sum_{i=1}^n \lvert y_i - y_m\rvert},
\end{equation}
$y_i^*$ is the actual target value of the i-th observation, $y_i$ is the predicted target value of the i-th observation, and $y_m$ is the mean of all actual target values.

\paragraph{Baseline Methods.}
We compare our method with 5 widely used feature generation methods, as well as random generation and feature dimension reduction methods: (1) \textbf{Base}: using the original dataset without feature generation. (2) \textbf{GRFG} \cite{wang2022group}:  iteratively generates new features and reconstructs an interpretable feature space through group-group interactions. (3) \textbf{DFS} \cite{kanter2015deep}: an expansion-reduction method that first expands and then selects feature, automatically generated features for the dataset. (4) \textbf{DIFER} \cite{difer2022}: Performs automated feature engineering in a continuous vector space, introducing an encoder-predictor-decoder to optimize features. It optimizes embeddings along the gradient direction and recovers improved features from the optimized embeddings. (5) \textbf{OpenFE} \cite{openfe2023}: Offers expert-level automated feature generation by integrating novel enhancement methods and a two-stage pruning algorithm, effectively identifying significant features.

\paragraph{Hyperparameter Settings.}
We utilize the Adam optimizer \cite{adam} to optimize the PPO algorithm, with a learning rate set to \(1 \times 10^{-4}\). The clipping parameter \(\epsilon\) for PPO is set to 0.2, and an entropy coefficient of \(1 \times 10^{-4}\) was applied to promote exploration. The number of PPO iterations is 10.
The split ratios for the training set, validation set, and test set are 0.6:0.2:0.2.
The encoder-decoder model incorporated 8 attention heads, with an embedding vector dimension of 128 and a model hidden layer dimension of 128. The maximum sequence length for feature transformations was set to 100. During pre-training, a temperature scaling parameter \(T\) ranging from 0.1 to 1 is used to balance exploration and exploitation.

\subsection{Overall Comparison}
In this experiment, we compare the performance of TSFG and baseline models for feature transformation. We evaluate classification tasks using \textit{F1-score}, accuracy, and precision, and regression tasks using 1-RAE and $R^2$. Table \ref{tab:1} shows the comparison results. We can see that in most cases, TSFG performs the best. By integrating the pre-training and fine-tuning stages, TSFG can accurately capture the internal patterns of the features, thereby identifying the optimal feature space. In conclusion, this experiment demonstrates the effectiveness of TSFG in feature transformation.

\subsection{Ablation Study}
To evaluate the contribution of each component in our framework, we perform ablation experiments. This experiment aims to verify whether each component of our method indeed has a positive impact on the final results. Therefore, we have developed two variants: (1) No Pre-Training: The model is trained without the pre-training phase. This variant is referred to as ``TSFG+".
(2) No PPO Fine-Tuning: The model is evaluated without reinforcement learning fine-tuning. This variant is referred to as ``TSFG\#".

\begin{figure}[ht]
    \centering
    \begin{subfigure}[b]{0.2\textwidth}
        \centering
        \includegraphics[width=\textwidth]{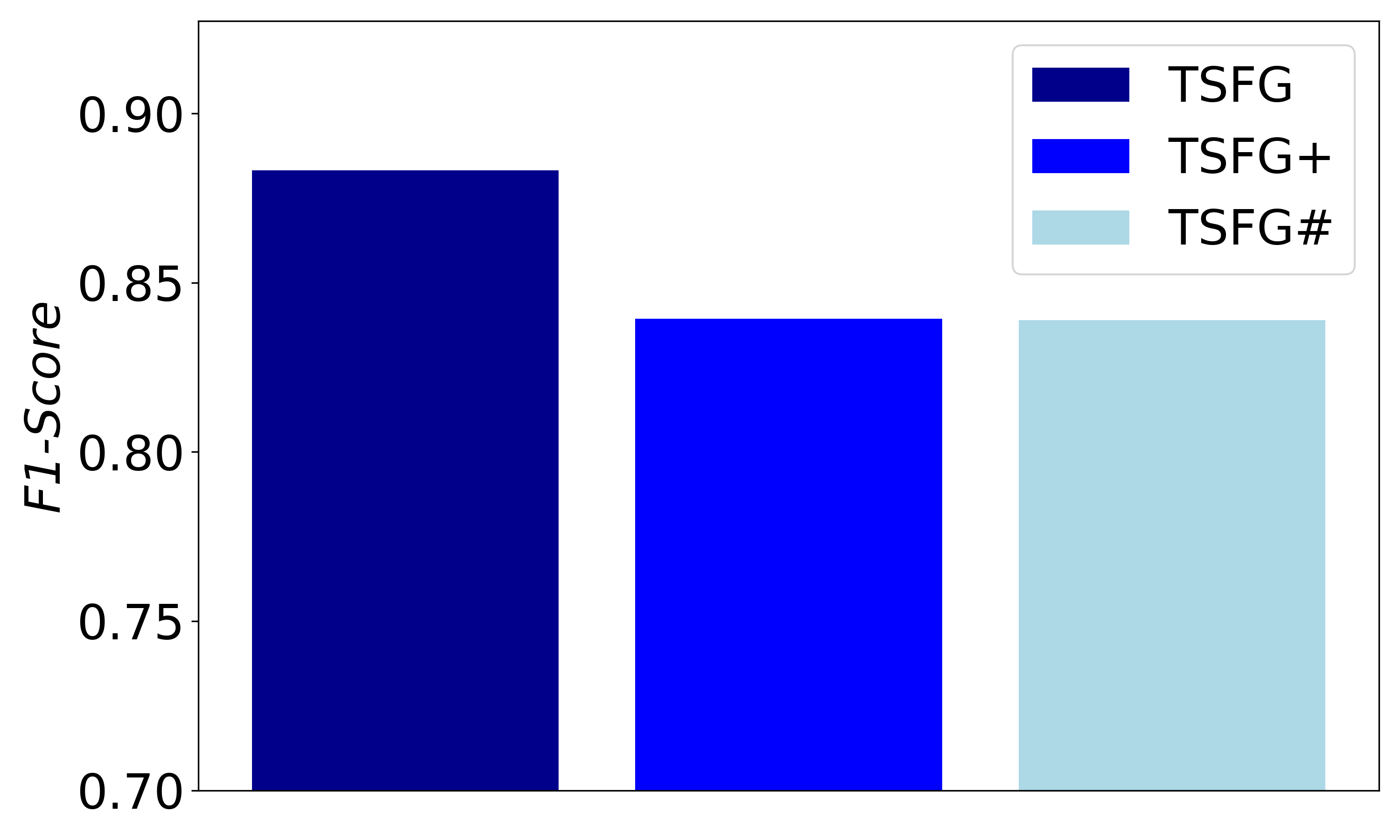}
        \caption{australian}
        \label{fig:Australian}
    \end{subfigure}
    \hfill
    \begin{subfigure}[b]{0.2\textwidth}
        \centering
        \includegraphics[width=\textwidth]{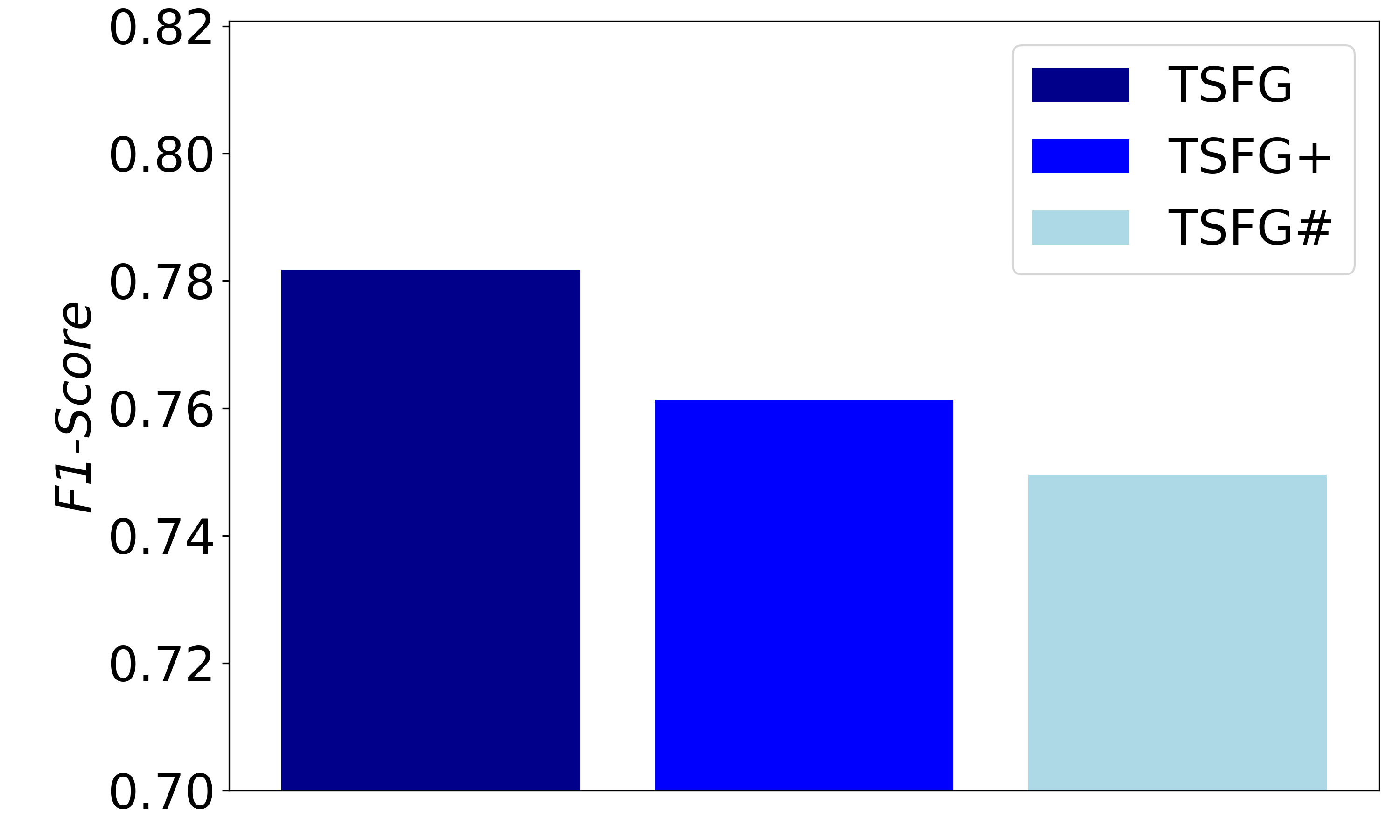}
        \caption{credit\_g}
        \label{fig:credit_g}
    \end{subfigure}

    \vspace{0.5em} 
    \begin{subfigure}[b]{0.2\textwidth}
        \centering
        \includegraphics[width=\textwidth]{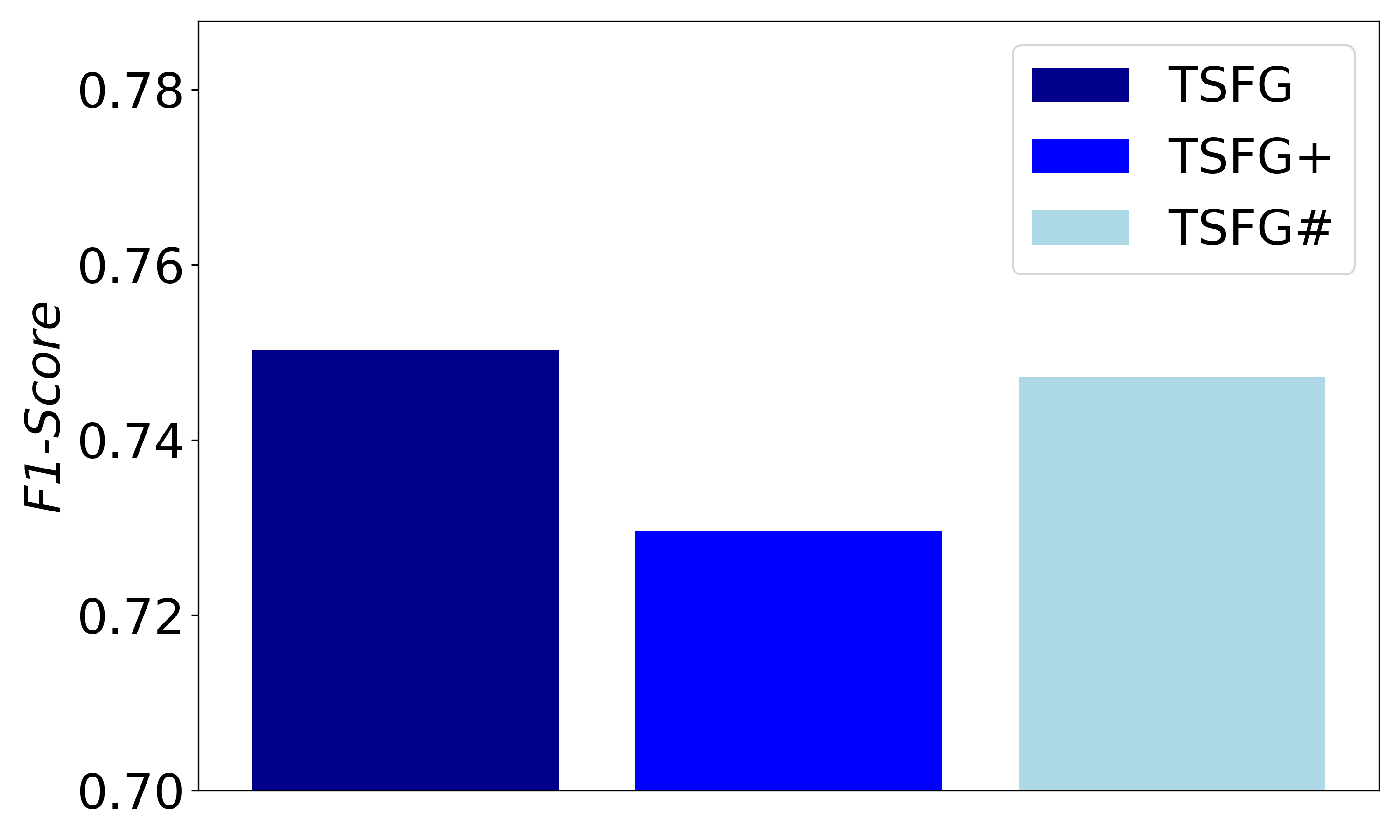}
        \caption{diabetes}
        \label{fig:diabetes}
    \end{subfigure}
    \hfill
    \begin{subfigure}[b]{0.2\textwidth}
        \centering
        \includegraphics[width=\textwidth]{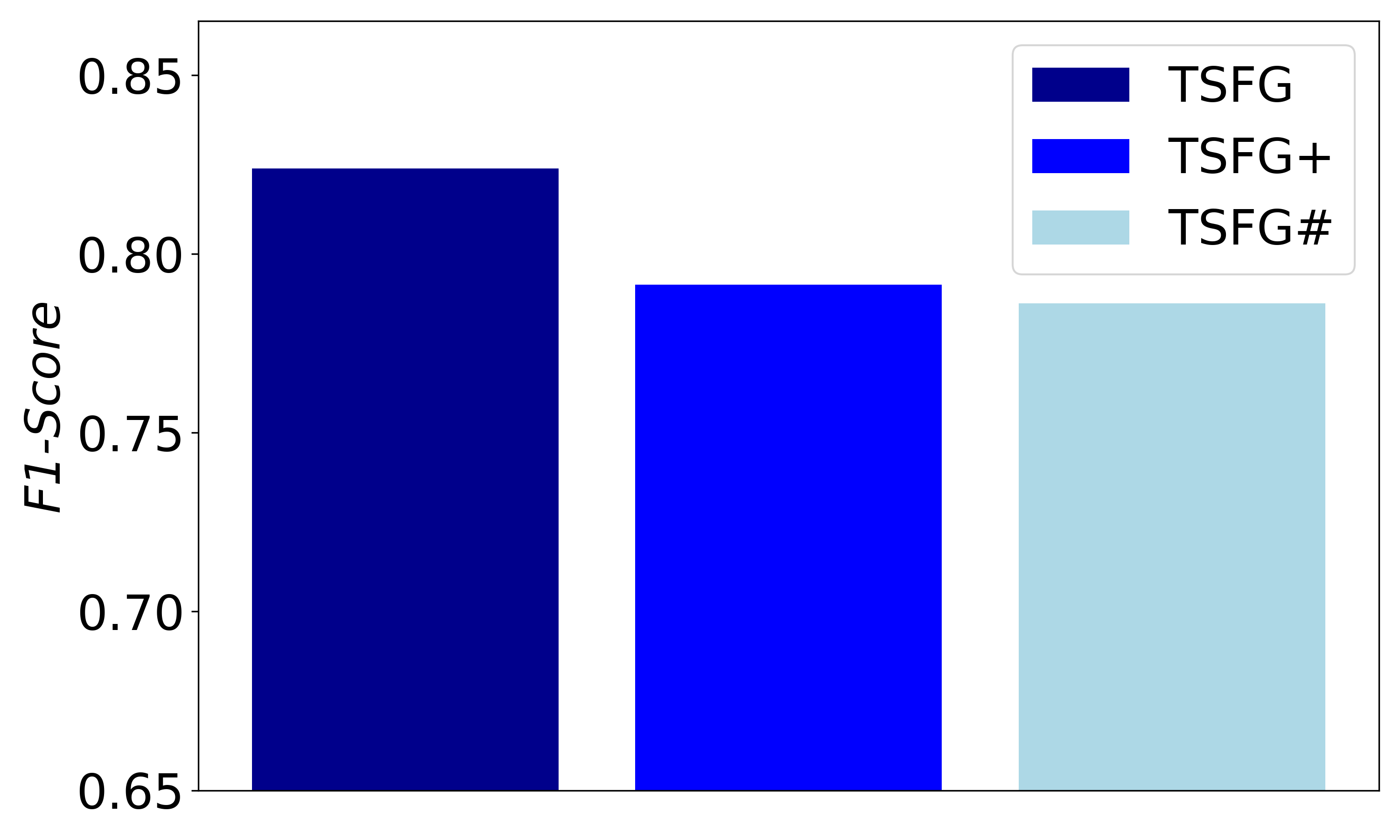}
        \caption{f5}
        \label{fig:f5}
    \end{subfigure}

    \caption{Results of ablation studies on different datasets.}
    \label{fig:ablation}
\end{figure}

As shown in Figure \ref{fig:ablation}, the results indicate that pre-training and PPO fine-tuning contribute significantly to the overall performance.

\subsection{Runtime Comparison}
As shown in Table \ref{tab:runtime_comparison}, we compare the runtime efficiency of the proposed framework with baseline methods. Despite incorporating reinforcement learning, our method achieves competitive runtime efficiency due to the use of pretraining. Compared with GRFG and DIFER, both of which use deep learning methods, the proposed method strikes a balance between computational cost and performance improvement.

\begin{table}[ht]
\centering
\renewcommand{\arraystretch}{1.0} 
\setlength{\tabcolsep}{4pt} 
\begin{tabular}{lccccc} 
\toprule
\textbf{Datasets} & \textbf{GRFG} & \textbf{DFS} & \textbf{OpenFE} & \textbf{DIFER} & \textbf{TSFG} \\ 
\midrule
australian & 458  & 1 & 10 & 1310 & 92 \\ 
credit\_g  & 598  & 1 & 25 & 1335 & 114 \\ 
diabetes   & 315  & 1 & 10 & 1950 & 104 \\ 
f5         & 1655 & 5 & 55 & 3567 & 129 \\ 
hepatitis  & 575  & 1 & 8  & 4000 & 114 \\ 
ionosphere & 1192 & 2 & 8  & 2217 & 130 \\ 
NPHA       & 704  & 2 & 9  & 1573 & 53 \\ 
Openml\_582 & 1533 & 7 & 10 & 4615 & 70 \\ 
Openml\_637 & 2657 & 2 & 11 & 7136 & 102 \\ 
Openml\_639 & 553  & 2 & 19 & 2774 & 73 \\ 
PimaIndian & 306  & 2 & 9  & 2103 & 103 \\ 
seismic    & 704  & 2 & 14 & 2160 & 139 \\ 
\bottomrule
\end{tabular}
\caption{Time cost comparisons with baselines, in seconds.} 
\label{tab:runtime_comparison} 
\end{table}

\subsection{Comparison on Different Downstream Tasks}

To evaluate the adaptability and effectiveness of the proposed framework, we compare its performance across different downstream machine learning models, including Random Forest, XGBoost, Support Vector Machines (SVM), and CatBoost (CAT). The experiment is conducted on the diabetes dataset.

As shown in Table \ref{tab:performance}, the proposed framework performs well across different downstream models. This highlights the framework's ability to generate high-quality features whose effectiveness is not dependent on the complexity of the task or the choice of downstream model. Furthermore, it demonstrates the framework's adaptability to various modeling requirements and its potential for broad applications in the field of feature generation.

\begin{table}[h]
\centering
\begin{tabular}{l|cccc}
\toprule
 & RF & XGB & SVM & CAT \\
\midrule
Base & 0.7062 & 0.7276 & 0.6829 & 0.7469 \\
GRFG & 0.7115 & 0.7478 & 0.7059 & 0.7387 \\
DFS & 0.6945 & 0.755 & 0.6603 & 0.7335 \\
OpenFE & 0.7233 & 0.743 & 0.697 & 0.7115 \\
DIFER & 0.7179 & 0.7306 & 0.6893 & 0.7658 \\
TSFG & 0.7602 & 0.753 & 0.7275 & 0.7693 \\
\bottomrule
\end{tabular}
\caption{Performance comparison of different downstream models.}
\label{tab:performance}
\end{table}

\subsection{Effectiveness of Generated Features}

\begin{figure}[ht]
    \centering
    \begin{subfigure}[b]{0.2\textwidth}
        \centering
        \includegraphics[width=\textwidth]{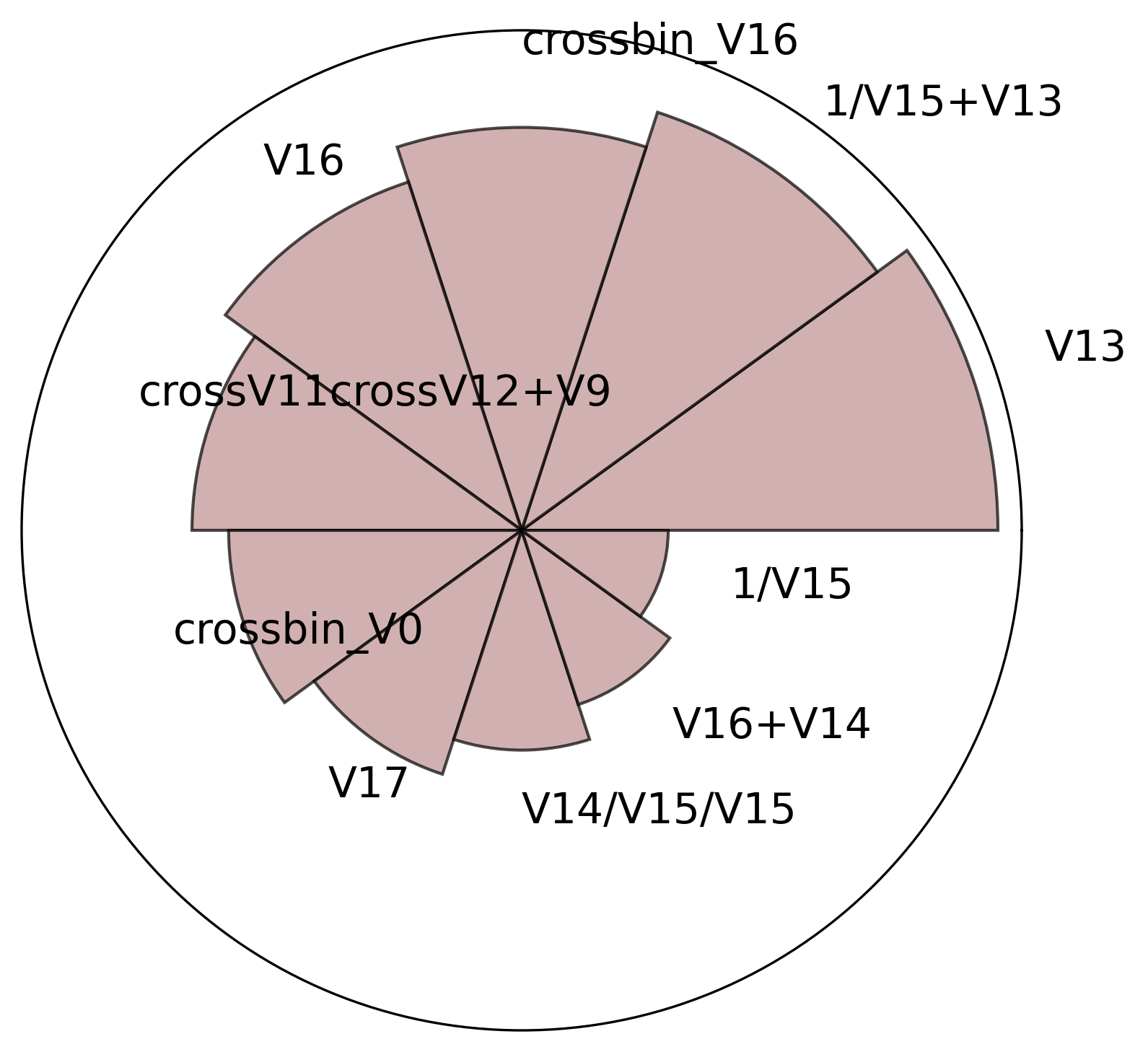}
        \caption{hepapits}
        \label{fig:feature_hepapits}
    \end{subfigure}
    \hfill
    \begin{subfigure}[b]{0.2\textwidth}
        \centering
        \includegraphics[width=\textwidth]{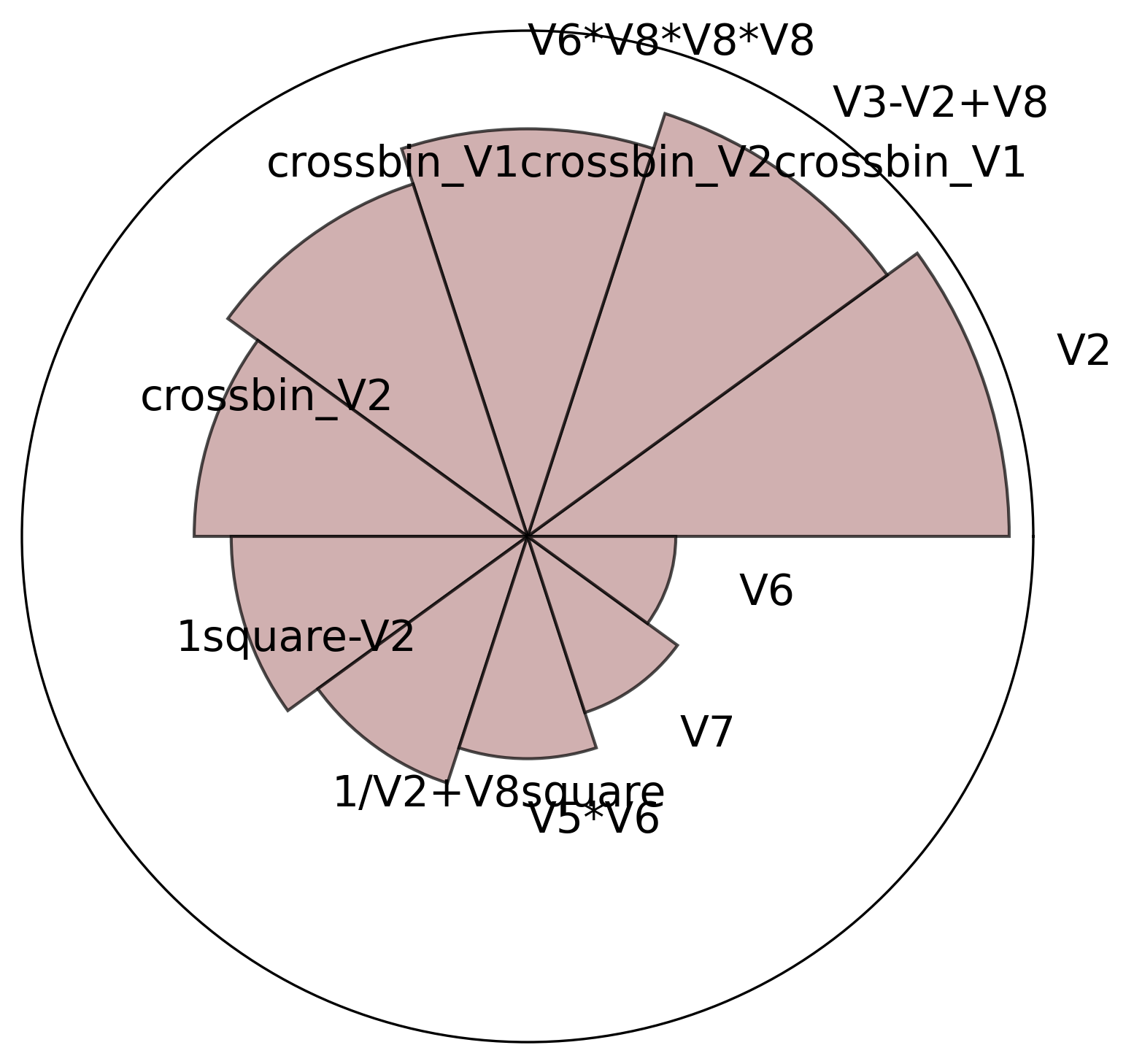}
        \caption{credit\_g}
        \label{fig:feature_credit}
    \end{subfigure}
    
    \caption{Feature importance analysis on different datasets.}
    \label{fig:feature_importance}
\end{figure}

We evaluate the quality of the generated features by analyzing their importance in downstream models. Feature importance scores from tree-based models (e.g., Random Forest) are used to measure the contribution of generated features. Figure \ref{fig:feature_importance} shows that the generated features have high importance scores, often surpassing original features, indicating their effectiveness in capturing complex relationships in data.

\section{Conclusion}
In this paper, we propose a novel automated feature generation framework that integrates reinforcement learning with a Transformer-based encoder-decoder architecture. The framework addresses the challenges of feature redundancy, inefficient exploration of the feature space, and adaptability to diverse datasets and tasks. Our method dynamically generates high-quality features, enhancing the performance of downstream machine learning models. Extensive experiments on various datasets validate the effectiveness of the proposed framework. The framework provides a solid foundation for advancing automated feature generation.

\section{Acknowledgments}
This work was supported by the Science Foundation of Jilin Province of China under Grant YDZJ202501ZYTS286, and in part by Changchun Science and Technology Bureau Project under Grant 23YQ05.

\bibliographystyle{named}
\bibliography{ijcai25}

\end{document}